\documentclass[letterpaper]{article} 
\usepackage{aaai24}  
\usepackage{times}  
\usepackage{helvet}  
\usepackage{courier}  
\usepackage[hyphens]{url}  
\usepackage{kotex}
\usepackage{verbatim}
\usepackage{float}
\usepackage{graphicx} 
\usepackage{booktabs}
\def\UrlFont{\rm}  
\usepackage{natbib}  
\usepackage{caption} 
\usepackage{algorithm}
\usepackage{algorithmic}

\usepackage{amssymb}
\usepackage{pifont}

\usepackage{newfloat}
\usepackage{listings}
\DeclareCaptionStyle{ruled}{labelfont=normalfont,labelsep=colon,strut=off} 
\floatstyle{ruled}
\newfloat{listing}{tb}{lst}{}
\floatname{listing}{Listing}
\usepackage{subcaption}

\title{Cross-Modal Retrieval Meets Inference:\\Improving Zero-Shot Classification with Cross-Modal Retrieval}
\author{
    Seongha Eom\textsuperscript{\rm 1},
    Namgyu Ho\textsuperscript{\rm 1},
    Jaehoon Oh\textsuperscript{\rm 2}\equalcontrib,
    Se-Young Yun\textsuperscript{\rm 1}\equalcontrib
}
\affiliations{
    \textsuperscript{\rm 1}Graduate School of AI, KAIST\\
    \textsuperscript{\rm 2}Samsung Advanced Institute of Technology \\
    \{doubleb, itsnamgyu, yunseyoung\}@kaist.ac.kr, jh0104.oh@gmail.com
}

\begin{document}

\maketitle

\begin{abstract}
Contrastive language-image pre-training (CLIP) has demonstrated remarkable zero-shot classification ability, namely image classification using novel text labels. Existing works have attempted to enhance CLIP by fine-tuning on downstream tasks, but these have inadvertently led to performance degradation on unseen classes, thus harming zero-shot generalization. This paper aims to address this challenge by leveraging readily available image-text pairs from an external dataset for cross-modal guidance \textit{during inference}. To this end, we propose X-MoRe, a novel inference method comprising two key steps: (1) cross-modal retrieval and (2) modal-confidence-based ensemble. Given a query image, we harness the power of CLIP's cross-modal representations to retrieve relevant textual information from an external image-text pair dataset. Then, we assign higher weights to the more reliable modality between the original query image and retrieved text, contributing to the final prediction. X-MoRe demonstrates robust performance across a diverse set of tasks without the need for additional training, showcasing the effectiveness of utilizing cross-modal features to maximize CLIP's zero-shot ability.
\end{abstract}

\section{Introduction}\label{sec:intro}

Pre-trained vision-language models, such as Contrastive Language-Image Pre-training (CLIP) \citep{radford2021learning}, have demonstrated  their effectiveness in acquiring transferable features from paired image and text data. The training process of CLIP harnesses two scalable elements: data and computational power. Firstly, the abundance of extensive image-text pairs available for training greatly contributes to its comprehensive learning. Secondly, CLIP's utilization of co-embeddings for images and language allows it to efficiently scale with computational resources. Consequently, CLIP has served as an alternative pre-training method, often replacing other visual pre-training approaches such as SimCLR \citep{chen2020simple} or MAE (Masked Autoencoder) \citep{he2022masked} across a wide range of applications. Subsequent techniques built upon language-image pre-training, like FLIP \citep{li2023scaling}, Audioclip \citep{guzhov2022audioclip} and Videoclip \citep{xu2021videoclip}, also show comparable favorable scaling results.

CLIP has been demonstrated to learn visual-language representations with substantial potential in zero-shot classification tasks. In order to enhance its effectiveness in downstream tasks, recent research studies have proposed the incorporation of additional learnable elements into CLIP, accompanied by adjustments made through the utilization of few-shot training datasets. \citet{zhou2022conditional, zhou2022learning} implement prompt tuning, while \citet{zhang2021tip} introduce adapters inspired by large language models. As a result, CLIP's pre-trained weights are kept fixed; instead, adaptable inputs or lightweight adapters are introduced to fine-tune textual or visual features.



Even though the inclusion of learnable inputs \citep{zhou2022conditional, zhou2022learning} or adapters \citep{zhang2021tip} demonstrates remarkable enhancements when dealing with seen classes during fine-tuning, their zero-shot performance on unseen classes tends to diminish. Thus, recent studies have indicated that these learnable inputs, specifically prompts, can be effectively tuned for a query image \emph{during inference}, eliminating the requirement for fine-tuning CLIP \citep{shu2022test, gao2022visual}. Meanwhile, \citet {guo2023calip} proposed using intermediate visual and textual features instead of learnable parameters for zero-shot enhancement of CLIP.

In addition, while these endeavors concentrate on leveraging visual input data to make well-founded model decisions, it is worth noting that the role of the language modality remain relatively unexplored for zero-shot classification. For instance, the text encoder within CLIP has been validated as beneficial for enhancing tasks in relation to other modalities \cite{chen2023difference}. \citet{kim2022diffusionclip} proved that text guidance from CLIP can give support for image generation task. \citet{menon2022visual} showed that visual descriptions of each category, instead of name, is helpful for explainable image classification task.

\begin{figure*}[!t]
    \centering
    \includegraphics[width=\linewidth]{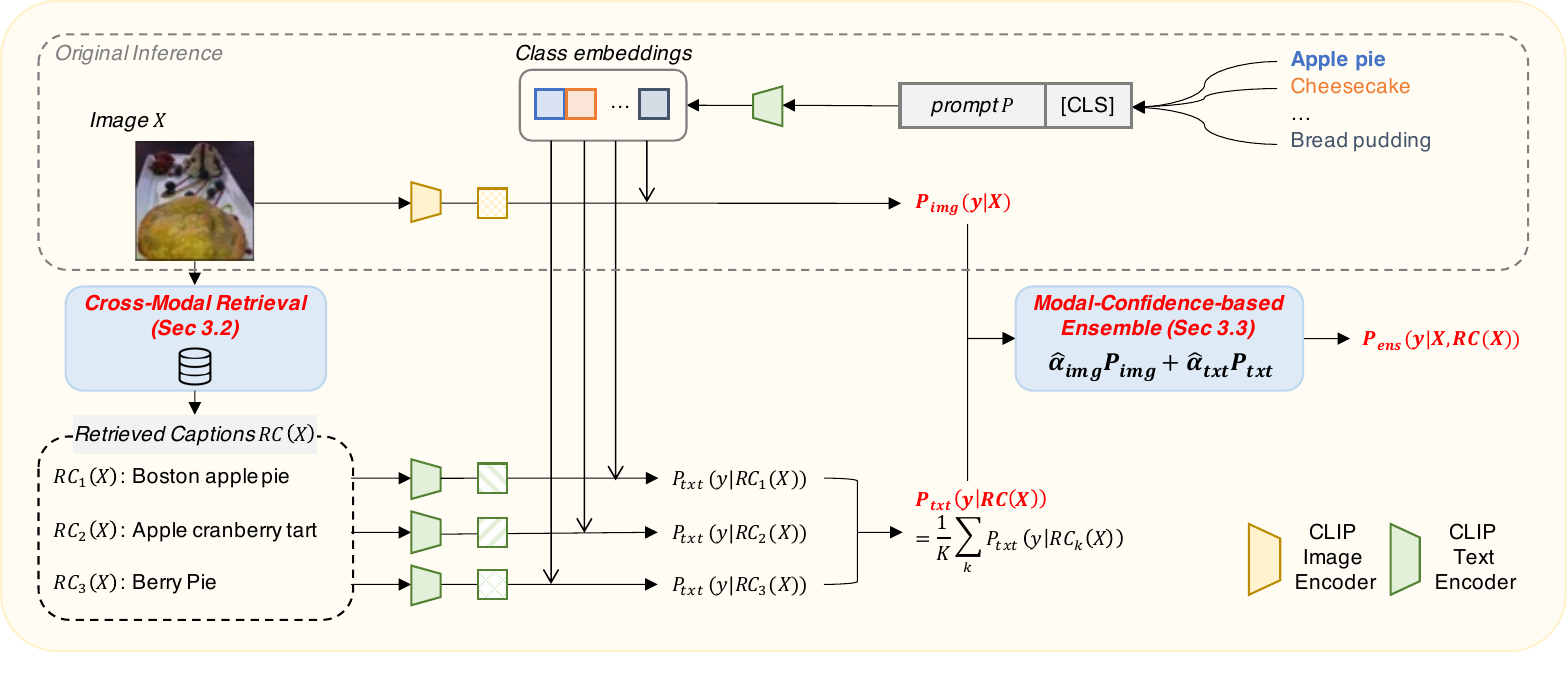}
    \caption{\texttt{X-MoRe} overview. Our pipeline first retrieves $K$ (e.g., $K$=3 in this figure) captions from an external dataset in cross-modal manner (Section 3.2). Then, we ensemble the image-modal prediction $P_{img}(y|X)$ and the text-modal prediction $P_{txt}(y|RC(X))$ based on each modal-confidence (Section 3.3). Note that the original inference uses only image-modal prediction $P_{img}(y|X)$ as shown in grey dotted line.
    }
    \label{fig:pipeline}
\end{figure*}

Therefore, to leverage the potential of the text modality during inference, we propose a novel \emph{tuning-free inference} method while preserving CLIP's zero-shot ability, called \texttt{X-MoRe} (\textbf{Cross}-\textbf{Mo}dal \textbf{Re}trieval based Inference). Our method, illustrated in Figure \ref{fig:pipeline}, is composed of two key steps: (1) cross-modal retrieval and (2) modal-confidence-based ensemble. In the cross-modal retrieval step, we create a pool of image-text pairs from an external dataset based on the similarity between a query image and images. We then meticulously retrieve texts from this pool of image-text pairs based on the direct similarity between a query image and texts. In the modal-confidence-based ensemble step, we ensemble both the image-modal and text-modal predictions using modal-confidence weighting.

Our work can be summarized as follows:

\begin{itemize}
    \item We propose \texttt{X-MoRe}, a novel tuning-free method for inference based on cross-modal retrieval. To capitalize on not only the image modality but also the text modality, \texttt{X-MoRe} consists of cross-modal retrieval and modal-confidence-based ensemble steps \textbf{(Section 3)}.
    \item We establish the superiority of our approach in zero-shot classification across 10 datasets and domain generalization tasks \textbf{(Section 4)}. Furthermore, we conduct several ablation studies to investigate the effectiveness of individual components of \texttt{X-MoRe} \textbf{(Section 5)}.
    \item We analyze the confidence of the image and text modalities and demonstrate that the correct modality exhibits higher confidence. Specifically, when the prediction from the image is correct and the prediction from the text is incorrect, the image modality is given higher priority, and vice versa for the text modality \textbf{(Section 6)}.
\end{itemize}

\section{Related Work}
\subsection{Vision Language Model}
Vision-Language Models (VLMs) are a class of models that integrate both visual and textual information to perform a wide range of tasks that require a deep understanding of both domains. Training VLMs typically involves contrastive learning, a self-supervised learning paradigm that encourages the model to learn representations by comparing positive and negative pairs.
The objective of VLMs is to align image and text representations in a shared embedding space, such that semantically similar images and texts are brought closer together while dissimilar ones are pushed apart.
Subsequently, to employ the trained model on downstream classification tasks, CLIP constructs the textual inputs by the category names and converts the original classification task into an image-text matching problem.
Consequently, CLIP has demonstrated its capacity to execute zero-shot recognition in open-vocabulary scenarios, yielding encouraging results on a range of benchmarks.

Various approaches have been developed to improve the classification performance of VLMs.
CLIP-Adapter \cite{gao2022visual} and Tip-Adapter \cite{zhang2021tip} use few-shot data to fine-tune lightweight adapters of CLIP on downstream tasks.
Inspired by prefix-tuning methods from language models \cite{li2021prefix}, Context Optimization (CoOp) \cite{zhou2022learning} has initiated a line of work that aims to preserve the original model and optimize context embeddings, i.e. surrounding the class label.
Recently, CALIP \cite{guo2023calip} introduced an approach that uses image-patchwise attention to enable fine-grained cross-modal alignment to enhance zero-shot classification in a parameter-free manner, extendable to few-shot settings.

\subsection{Cross-modal Retrieval}
Retrieval based module is also adopted for natural language generation in KNN-LM \cite{khandelwal2019generalization} and image recognition in \cite{iscen2023retrieval, iscen2023improving, hu2023reveal}.
Different from previous works which typically store features and label from models as key-value pair, we store image index and captions in natural language text as key-value pair which is much more lightweight.
In another line of work, information is retrieved from knowledge graphs to improve the factuality of language models in applications such as dialogue and question answering \cite{kang2023knowledge, baek2023knowledge} or aid language models in tasks which require domain-specific knowledge \cite{kang2022kala}.
Inspired by the success of knowledge fusion from cross-modal retrieval, we collect relevant captions for the query image by performing KNN search over an external image-text pair dataset.


\subsection{Cross-modal Ensemble}

Cross-modal ensembles have been employed to solve the multilingual video corpus moment retrieval (mVCMR) task \cite{liu2022cross, lei2021mtvr}.
This task involves retrieving short moments from a large corpus of videos with subtitles, given a natural language query, in various languages.
mVCMR methods ensemble information from the subtitles and the visual content to determine the similarity of video moments to a given query.
Meanwhile, recent work from the CLIP literature utilizes an ensemble of various visual information for test-time prompt-tuning \cite{shu2022test}.
In particular, the prompt is optimized to perform consistently across multiple augmented views of each query image.
To avoid noisy augmentations, the authors introduce confidence selection, which discards views with low-entropy class predictions.
Our method can be seen as a form of cross-modal ensemble, utilizing information from the query image as well as retrieved captions.
Inspired by \citet{shu2022test}, we employ entropy to determine ensemble weights for each modality, to enhance the reliability of predictions.


\section{X-MoRe: Cross-Modal Retrieval Inference}

In this section, we propose \texttt{X-MoRe}, which is a novel inference method based on cross-modal retrieval and modal-confidence-based ensemble.

\subsection{Prerequisite: CLIP Inference}

CLIP comprises an image encoder, denoted as $f$, and a text encoder, denoted as $g$, as its core components. They respectively embed images and textual descriptions into high-dimensional representations. During CLIP training, it learns to align these image and text representations in a shared embedding space. When CLIP is applied to a zero-shot classification task, a query image $X$ is embedded using $f$ and the text labels with a pre-defined prompt (e.g., ``a photo of a [CLS]'') are embedded using $g$. In the image-text embedding space, a query image has similarities measured with all of $C$ text labels. Here, we define the \emph{image-modal probability} $P_{img}(y|X)$ from the similarities. Then, a query image is predicted as the text label with the highest probability. This zero-shot classification inference is illustrated with a dotted gray line in Figure \ref{fig:pipeline}.


\subsection{Cross-modal Retrieval}\label{subsec:xmore}

\begin{figure}[t!]
    \centering
    \includegraphics[width=0.9\columnwidth]{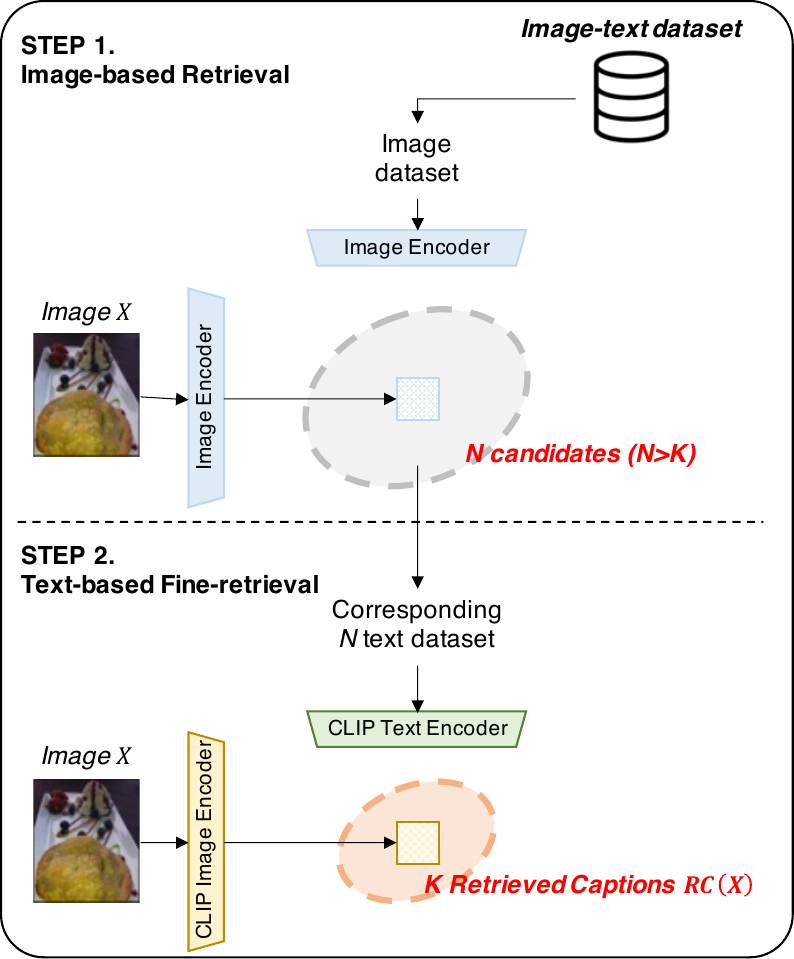}
    \caption{An overview of our two-stage cross-modal retrieval, starting with image-based retrieval and followed by text-based fine-retrieval. Given a query image $X$, we first obtain $N$ image-text candidates from an external image-text dataset based on image similarity. This process is then refined to retrieve $K$ texts by directly comparing the query image $X$ to the texts of the $N$ candidates.}
    \label{fig:retrieval}
\end{figure}

We propose a two-step cross-modal retrieval method: image-based retrieval then text-based fine-retrieval. Figure \ref{fig:retrieval} illustrates the retrieval process. Given a query image $X$, we coarsely obtain $N$ images and corresponding texts based on the similarity between a query image $X$ and the images from an external image-text dataset (e.g., public dataset or web-scale image-text database). Then, we finely obtain $K$ texts based on the similarity between a query image $X$ and texts from $N$ image-text candidates. This stage is designed to be independent of inference, thereby reinforcing retrieval capabilities through the utilization of enhanced language-image pre-trained models.

\subsubsection{STEP1. Image-based Retrieval}
We initially construct a set of $N$ image-text candidates for a given query image $X$ by focusing on the image modality. This is achieved through a KNN retrieval module, which searches the nearest neighbors from an external image-text dataset, following previous works \citep{iscen2023improving, iscen2023retrieval, hu2023reveal, khandelwal2019generalization}. For instance, \citet{iscen2023retrieval} employ the image modality to search the nearest neighbors and retrieve the corresponding texts from image-text pairs. The key difference from prior works is that we establish a set of $N$ candidates larger than $K$, because the text modality is not directly considered within this step.


\subsubsection{STEP2. Text-based Fine-retrieval}
In order to retrieve textual content associated with a query image $X$ more accurately, we fine-retrieve $K$ text captions $RC(X)$ from the pool of coarsely retrieved $N$ image-text candidates. This procedure is important because there are cases where the ranking of similarity between a query image $X$ and images does not align with the ranking of similarity between the query image and their corresponding texts. Put simply, in the cross-modal embedding space, even if image $X$ is closer to ${img}_1$ than to ${img}_2$, it could be closer to ${txt}_2$ (the corresponding text of ${img}_2$) than to ${txt}_1$ (the corresponding text of ${img}_1$). Without this direct approach, it is hard to conceive that the retrieving texts used for predicting images are well describing the inputs.

After fine-retrieval, we define the \emph{text-modal probability} $P_{txt}(y|RC(X))$ by averaging predictions obtained from the $K$ retrieved captions $RC(X)$. This aggregation is utilized to create a single representative distribution for the text modality, thereby aligning it with the image-modal probability $P_{img}(y|X)$. Additionally, it prevents the text-modal probability from making overconfident incorrect predictions, which will be addressed in Section \ref{subsec:text_lower}.



\subsection{Modal-confidence-based Ensemble}\label{sec:modal_conf}

\begin{figure}[!t]
    \centering
    \includegraphics[width=\columnwidth]{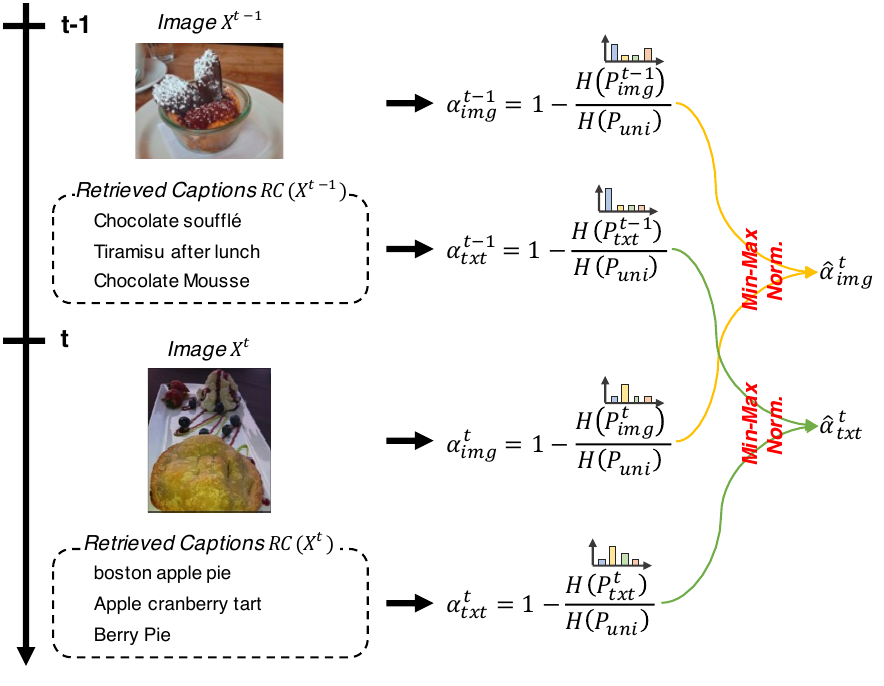}
    \caption{An example of calculating modal-confidences at time $t$. We first compute modal-confidences (i.e., $\alpha^t_{img}$ and $\alpha^t_{txt}$) based on entropy of each prediction probability. Then, we independently calibrate these modal-confidences using min-max normalization, resulting in adjusted confidences (i.e., $\widehat\alpha^t_{img}$ and $\widehat\alpha^t_{txt}$), within the confidences encountered from previous test time.}
    \label{fig:weighted_ensemble}
\end{figure}

After acquiring the image-modal probability $P_{img}(y|X)$ through the original inference and the text-modal probability $P_{txt}(y|RC(X))$ through the cross-modal retrieved inference, the two probabilities are ensembled based on modal-confidence for the \emph{cross-modal probability} $P_{ens}(y|X,RC(X))$, which is the final prediction of \texttt{X-MoRe}.

For our modal-confidence, we contemplate a scenario where test images arrive sequentially, one by one. Indeed, when the test images come in a batch, if they are processed one by one, it is essentially similar to our scenario. Specifically, a query image $X$ and retrieved captions $RC(X)$ at test time $t$ is denoted as $X^t$ and $RC(X^t)$. Then, for simplicity, we express $P_{img}(y|X^t)$, $P_{txt}(y|RC(X^t))$, and $P_{ens}(y|X^t,RC(X^t))$ as $P^t_{img}$, $P^t_{txt}$, and $P^t_{ens}$, respectively.

Image-confidence $\alpha^t_{img}$ and text-confidence $\alpha^t_{txt}$ at time $t$ are fundamentally based on entropy as follows:

\begin{equation}
    \alpha^t_{img} = 1 - \frac{H(P^t_{img})}{H(P_{uni})}, \quad  \alpha^t_{txt} = 1 - \frac{H(P^t_{txt})}{H(P_{uni})}
\end{equation}

\noindent where $H(\cdot)$ is the entropy of a probability distribution and $P_{uni}$ is an uniform distribution over $C$ classes (i.e., $[\frac{1}{C}, \cdots, \frac{1}{C}]$). Modal-confidence values move closer to 0 when the probability distribution resembles a uniform distribution, whereas they move closer to 1 when the probability distribution concentrates on a specific class.

After calculating both confidence, we independently adjust them with min-max normalization within the confidences accumulated from previous test time. This is because the two different modalities have different entropy distributions. For instance, say that both of $\alpha^t_{img}$ and $\alpha^t_{txt}$ are 0.7. However, if the maximum value of $\alpha_{img}$ is 0.6 and the minimum value of $\alpha_{txt}$ is 0.8 until $t-1$, then $X^t$ is the most confident image and $RC(X^t)$ is the least confident text. With this concern, adjusted image-confidence $\widehat{\alpha}^t_{img}$ and adjusted text-confidence $\widehat{\alpha}^t_{txt}$ are as follows, depicted in Figure \ref{fig:weighted_ensemble}:

\begin{equation}
    \widehat{\alpha}^t_{img} = \frac{\alpha^t_{img} - \min(\alpha^{1:t}_{img})}{\max(\alpha^{1:t}_{img}) - \min(\alpha^{1:t}_{img})}
\end{equation}
\begin{equation}
    \widehat{\alpha}^t_{txt} = \frac{\alpha^t_{txt} - \min(\alpha^{1:t}_{txt})}{\max(\alpha^{1:t}_{txt}) - \min(\alpha^{1:t}_{txt})}
\end{equation}

\noindent where $\min(\alpha^{1:t})$ and $\max(\alpha^{1:t})$ can be updated at each time $t$. With these adjusted modal-confidences, the cross-modal probability $P^t_{ens}$ for the final prediction at test time $t$ is calculated as follows:
\begin{equation}
    P^t_{ens} = \widehat{\alpha}^t_{img}P^t_{img} + \widehat{\alpha}^t_{txt}P^t_{txt}
\end{equation}

Our modal-confidence does not represent the directly relative confidence between the modalities, but rather it represents the relative confidence in each individual modality. It means that both $\widehat{\alpha}^t_{img}$ and $\widehat{\alpha}^t_{txt}$ independently take values between 0 and 1. For instance, if $\widehat{\alpha}^t_{img}$ is 1, it means that the query image at time $t$ has the highest confidence (i.e., lowest entropy) among the images encountered so far; conversely, if it is 0, it means that the query image at time $t$ has the lowest confidence (i.e., highest entropy).

\section{Experiment}

\begin{table*}[t!]
\small
\centering
\setlength{\tabcolsep}{3pt}
    \begin{tabular}{l|cccccccccc|c}
    \toprule
    Method & Flower102 & DTD & Pets & Cars & UCF101 & Caltech101 & Food101 & SUN397 & Aircraft & EuroSAT & Avg. \\
    \midrule
    CLIP-RN50$^{\dagger}$ & 61.75 & 40.37 & 83.57 & 55.70 & 58.84 & 85.88 & 73.97 & 58.80 & 15.66 & 23.69 & 55.82 \\
    Caption Ensemble & 59.36 & 35.40 & 62.74 & 43.82 & 51.44 & 81.82 & 70.00 & 49.44 & 19.50 & 29.65 & 50.32 \\
    CALIP & 60.90 & 39.89 & 81.25 & 55.64 & 54.59 & 83.94 & 71.49 & 56.84 & 15.72 & 24.22 & 54.45\\
    TPT$^{\dagger}$ & 62.69 & 40.84 & 84.49 & 58.46 & 60.82 & 87.02 & 74.88 & 61.46 & 17.58 & 28.33 & 57.66 \\
    X-MoRe & \textbf{69.78} & \textbf{45.17} & \textbf{84.96} & \textbf{60.66} & \textbf{63.57} & \textbf{90.10} & \textbf{79.52} & \textbf{62.61} & \textbf{21.14} & \textbf{32.94} & \textbf{61.04}\\
    \midrule
    CLIP-ViT-B/16$^{\dagger}$ & 67.44 & 44.27 & \textbf{88.25} & 65.48 & 65.13 & 93.35 & 83.65 & 62.59 & 23.67 & 42.01 & 63.58 \\
    Caption Ensemble & 61.39 & 36.05 & 60.59 & 44.84 & 51.69 & 85.40 & 71.64 & 49.61 & 49.61 & 33.32 & 51.48\\
    CALIP & 67.64 & 44.44 & 87.82 & 65.80 & 64.05 & 93.27 & 82.76 & 62.52 & 24.12 & 42.27 & 63.47\\
    TPT$^{\dagger}$ & 68.98 & \textbf{47.75}& 87.79 & 66.87 & \textbf{68.04} & \textbf{94.16} & \textbf{84.67} & \textbf{65.50} & 24.78 & 42.44 & 65.10 \\
    X-MoRe & \textbf{71.65} & 47.36 & 87.61 & \textbf{67.35} & 66.56 & 93.91 & 84.54 & 65.32 & \textbf{25.74} & \textbf{45.81} & \textbf{65.57}\\
    \bottomrule
    \end{tabular}
    \caption{Zero-shot performance across image classification benchmark datasets. We consistently used 16 captions for caption ensemble or \texttt{X-MoRe}. $^{\dagger}$ performances are from \citet{shu2022test}.}
    \label{tab:comp_finegrained}
\end{table*}

\begin{table*}[t!]
\small
\centering
    \begin{tabular}{l|ccccc|cc}
    \toprule
    Method & ImageNet & ImageNet-A & ImageNet-V2 & ImageNet-R & ImageNet-Sketch & Avg. & Variants Avg. \\
    \midrule
    CLIP-RN50$^{\dagger}$ & 58.16 & 21.83 & 51.41 & 56.15 & 33.37 & 44.18 & 40.69 \\
    Caption Ensemble & 47.21 & 21.47 & 39.26 & 46.06 & 28.85 & 36.57 & 33.91\\
    CALIP  & 58.21 & 21.67 & 51.44 & 55.96 & 33.30 & 44.12 & 40.59\\
    TPT$^{\dagger}$ & 60.74 & \textbf{26.67} & 54.70 & 59.11 & 35.09 & 47.26 & 43.89 \\
    X-MoRe & \textbf{62.71} & 26.49 & \textbf{55.25} & \textbf{62.21} & \textbf{40.41} & \textbf{49.41} & \textbf{46.09}\\
    \midrule
    CLIP-ViT-B/16$^{\dagger}$ & 66.73 & 47.87 & 60.86 & 73.98 & 46.09 & 59.11 & 57.20 \\
    Caption Ensemble & 47.11 & 22.62 & 39.64 & 47.84 & 29.34 & 37.31 & 34.86 \\
    CALIP & 66.74 & 47.76 & 60.76 & 73.99 & 46.12 & 59.07 & 57.16 \\
    TPT$^{\dagger}$ & \textbf{68.98} & \textbf{54.77} & \textbf{63.45} & \textbf{77.06} & \textbf{47.94} & \textbf{62.44} & \textbf{60.81} \\
    X-MoRe & 67.01 & 46.56 & 59.87 & 73.60 & 46.61 & 58.73 & 56.66 \\
    \bottomrule
    \end{tabular}
    \caption{Zero-shot performance on ImageNet and its variants. We consistently used 16 captions for caption ensemble or \texttt{X-MoRe}. $^{\dagger}$ performances are from \citet{shu2022test}.}
    \label{tab:comp_imagenet}
\end{table*}


\subsection{Settings}
We evaluate our approach using two types of CLIP models: CLIP-RN50 ($f$ is ResNet50 \citep{he2016deep} and $g$ is 12-layer Transformer \citep{vaswani2017attention}) and CLIP-ViT-B/16 ($f$ is ViT-B/16 \citep{dosovitskiy2020image} and $g$ is 12-layer Transformer \citep{vaswani2017attention}). During inference, we resize all test images to a resolution of 224×224 and employ the general textual template ``a photo of a [CLS]'' for classification. Our baseline comparisons include both tuning-free inference (e.g., pre-trained CLIP $P_{img}$ \cite{radford2021learning}, an ensemble of retrieved captions $P_{txt}$ ($K$=16), and CALIP \cite{guo2023calip}) and tuning-based inference (e.g., TPT \citep{shu2022test}).

For the cross-modal retrieval in \texttt{X-MoRe}, we utilize laion-400m \cite{schuhmann2021laion} public dataset as an external image-text dataset. In the image-based retrieval step, we use visual embeddings extracted from an image encoder of CLIP-ViT-B/32, provided by the CLIP retrieval website\footnote{https://github.com/rom1504/clip-retrieval}. In the text-based fine-retrieval step, we use CLIP-ViT-L/14, discussed in Section \ref{subsec:img_encoder}. Across all experiments, we establish a pool of $128$ image-text candidates, and subsequently, fine-retrieve $16$ captions (i.e., $N$ is 128 and $K$ is 16), unless otherwise specified. We repeat experiments three times on different seeds.

\subsection{Zero-shot Classification}

\subsubsection{Datasets}
We evaluate zero-shot ability across 10 image classification datasets: Caltech101 \citep{fei2004learning}, OxfordPets \citep{parkhi2012cats}, StanfordCars \citep{krause20133d}, Flowers102 \citep{nilsback2008automated}, Food101 \citep{bossard2014food}, FGVCAircraft \citep{maji2013fine}, SUN397 \citep{xiao2010sun}, DTD \citep{cimpoi2014describing}, EuroSAT \citep{helber2019eurosat}, and UCF101 \citep{soomro2012ucf101}. These datasets cover a wide spectrum of tasks, including generic object, scene, action, and fine-grained classification, along with specialized tasks such as texture recognition and satellite image analysis.

\subsubsection{Results}
Table \ref{tab:comp_finegrained} provides the zero-shot performance results across 10 datasets. \texttt{X-MoRe} demonstrates the highest average accuracy in comparison to the baseline methods, even including a tuning-based inference approach (i.e., TPT). Specifically, for CLIP-RN50, our method consistently enhances the zero-shot performance across all datasets. It is notable that caption ensemble displays the lowest accuracy, implying the importance of ensemble with the image modality. This observation also highlights the effectiveness of \texttt{X-MoRe}'s balanced utilization of both modalities.


\subsection{Domain Generalization}

\subsubsection{Datasets}
For the evaluation of general domain generalization, ImageNet \cite{deng2009imagenet} is used as the source dataset and four variants of ImageNet are used as the target datasets (i.e., ImageNetV2 \cite{recht2019imagenet}, ImageNet-Sketch \cite{wang2019learning}, ImageNet-A \cite{hendrycks2021natural} and ImageNet-R \cite{hendrycks2021many}). However, our method and the baselines are not directly linked to the training approach. Thus, we adhere to the zero-shot evaluation protocol. In detail, ImageNetV2 is a recreated test set obtained from various sources using the same data collection process as ImageNet. ImageNet-Sketch contains sketches of the same 1,000 classes found in ImageNet. Meanwhile, ImageNet-A and ImageNet-R consist of 200 classes, a subset of ImageNet's 1,000 classes. ImageNet-A comprises real-world images that have been adversarially filtered, causing current ImageNet classifiers to perform poorly. ImageNet-R presents ImageNet classes in diverse visual styles like paintings, cartoons, and sculptures.


\subsubsection{Results}
Table \ref{tab:comp_imagenet} provides the performance evaluation on ImageNet and its variants. For CLIP-RN50, similar to zero-shot classification experiment, \texttt{X-MoRe} outperforms all other baselines, including the tuning-based inference approach. In fact, our approach can be viewed as an ensemble of `CLIP-RN50' and `Caption Ensemble' with modal-confidences, leading to performance enhancement. In other words, when the two modalities have comparable performance, our method offers significant performance improvement. However, for CLIP-ViT-B/16, our method decreases the performance. We speculate that, based on the results of large gap between `CLIP-ViT-B/16' and `Caption Ensemble,' our approach encounters challenges when one modality exhibits significant underperformance.

\section{Ablation Study}
In this section, we delve into ablation studies to provide further evidence for the effectiveness of our approach.



\subsection{Importance of Text-based Fine-retrieval}
We compare indirect and direct text retrieval to highlight the significance of text-based fine-retrieval. Indirect text retrieval indicates the process of searching $K$ corresponding texts through image-based retrieval. Table \ref{tab:abl2} demonstrates that directly searching by measuring the similarity between a query image $X$ and texts within a shared co-embedding space (i.e., text-based fine-retrieval in our algorithm) is one of the key factors.




\begin{table}[h!]
\centering
    \begin{tabular}{lcccc}
    \toprule
    Approach & Flower102 & Caltech101 \\ 
    \midrule 
    Indirect & 65.04 & 88.92 \\
    Direct (Ours) & \textbf{69.87} & \textbf{90.10} \\

    \bottomrule
    \end{tabular}
    \caption{Performance of zero-shot classification whether text retrieval is direct or not. CLIP-RN50 is used for inference.}
    \label{tab:abl2}
\end{table}

\subsection{Encoders for Text-based Fine-retrieval}\label{subsec:img_encoder}
The retrieval stage is intentionally structured to be distinct from inference to leverage the capabilities of improved language-image pre-trained models for cross-modal retrieval, as mentioned in Section \ref{subsec:xmore}. Specifically, for the text-based fine-retrieval, we can have the flexibility to scale the encoder. Table \ref{tab:abl1} reveals that using the largest image encoder with a smaller patch size enhances the retrieval quality, resulting in performance improvement. Therefore, CLIP-ViT-L/14 is used for our fine-retrieval.


\begin{table}[h]
\centering
    \begin{tabular}{lcccc}
    \toprule
    Image Encoder & Flower102 & Caltech101 \\
    \midrule
    ViT-B/32 & 64.84 & 88.88 \\
    ViT-B/16 & 66.46 & 89.21 \\
    ViT-L/14 & \textbf{69.87} & \textbf{90.10} \\
    \bottomrule
    \end{tabular}
    \caption{Performance of zero-shot classification according to the image encoder for the text-based fine-retrieval. CLIP-RN50 is used for inference.}
    \label{tab:abl1}
\end{table}

\subsection{Number of Retrieved Captions}\label{subsec:num_cap}
\begin{figure}[h]
\small
    \centering
    \begin{minipage}{.49\linewidth}
        \includegraphics[width=\linewidth]{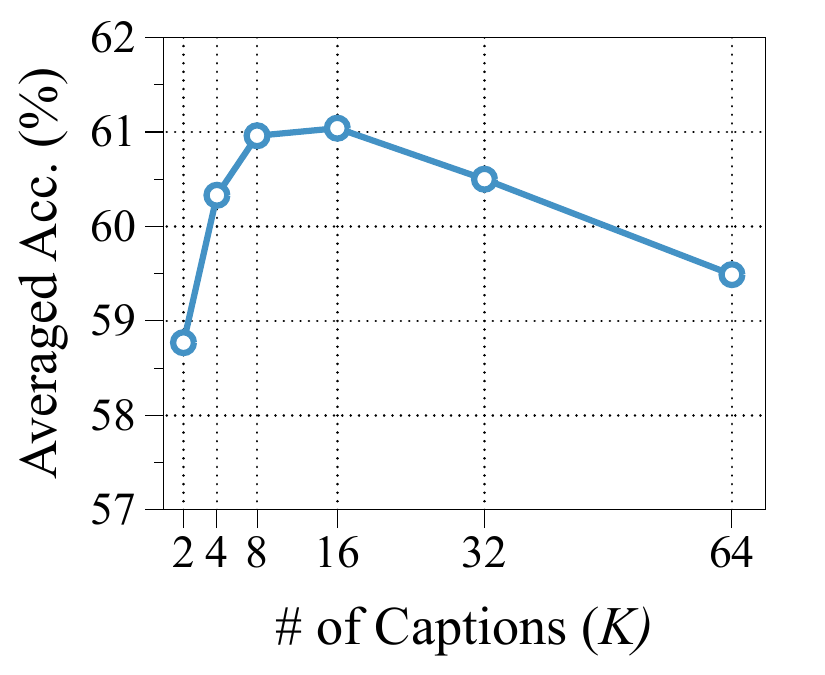}
        \subcaption{CLIP-RN50}
    \end{minipage}
    \hfill
    \begin{minipage}{.49\linewidth}
        \includegraphics[width=\linewidth]{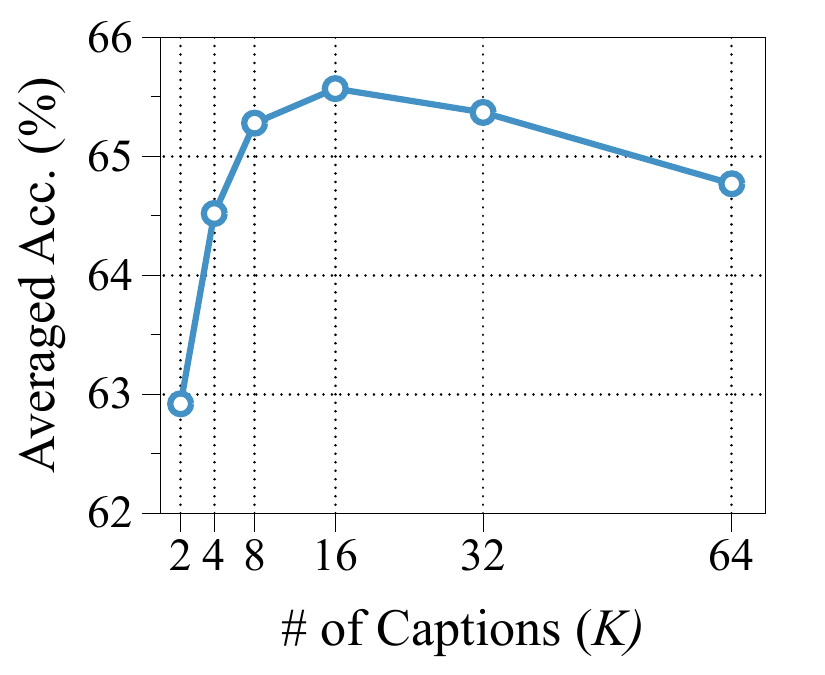}
        \subcaption{CLIP-ViT-B/16}
    \end{minipage}
    \caption{Averaged accuracy across 10 datasets according to the number of retrieved captions $K$.}
    \label{fig:variousK}
\end{figure}

Figure \ref{fig:variousK} illustrates the averaged zero-shot classification performance of \texttt{X-MoRe} across 10 datasets, varying the number of retrieved captions (i.e., $K$). The left and right figures depict the performance of CLIP-RN50 and CLIP-ViT-B/16, respectively. It is observed that the highest performance is achieved with 16 captions. This indicates that using an adequate number of retrieved texts is one of the key factors in achieving improved performance.

\subsection{Modal-confidence-based Ensemble}
Lastly, we examine the importance of modal-confidence-based ensemble. The adjusted modal-confidence values (i.e., $\widehat\alpha^t_{img}$ and $\widehat\alpha^t_{txt}$) approach 1, if these modalities are reliable. To assess this, we compare our approach with the equal-confidence strategy. This strategy assumes that the two modalities always have the same reliability (i.e., $\widehat\alpha^t_{img}=\widehat\alpha^t_{txt}=1$). Table \ref{tab:abl3} demonstrates that our modal-confidence-based ensemble slightly improves the performance in comparison to the straightforward equal-confidence-based ensemble.



\begin{table}[h!]
\setlength{\tabcolsep}{3pt}
\centering
    \begin{tabular}{lcccc}
    \toprule
    Approach & Flower102 & Caltech101 & Cars  & EuroSAT\\
    \midrule
    Equal-conf. & 69.63 & 90.10 & 60.41  & 32.18\\
    Modal-conf. & \textbf{69.87} & \textbf{91.15} & \textbf{60.65} & \textbf{32.94} \\
    \bottomrule
    \end{tabular}
    \caption{Performance of zero-shot classification whether the confidences of two modalities are equal (Equal-conf.) or not (Modal-conf.). CLIP-RN50 is used for inference.}
    \label{tab:abl3}
\end{table}

\begin{table*}[t!]
\small
\centering
\setlength{\tabcolsep}{3pt}
    \begin{tabular}{lcccccccccc}
    \toprule
    Modality & Flower102 & DTD & Pets & Cars & UCF101 & Caltech101 & Food101 & SUN397 & Aircraft & EuroSAT \\
    \midrule
    Image & 1.37 & 1.88 & 0.64 & 1.57 & 1.52 & 0.69 & 1.00 & 1.77 & 2.91 & 
1.75 \\
    Text ($K$=1)  &  0.32 & 1.08 & 0.57 & 0.52 & 0.68 & 0.40 & 0.29 & 1.10 & 1.99 & 1.32\\
    Text ($K$=8)  & 0.96 & 1.73 & 1.16 & 1.34 & 1.35 & 0.68 & 0.82 & 1.74 & 2.53 & 1.79 \\
    Text ($K$=16)  & 1.16 & 1.87 & 1.30 & 1.52 & 1.53 & 0.73 & 1.01 & 1.89 & 2.64 & 1.85 \\
    Text ($K$=64)  & 1.75 & 2.18 & 1.62 & 2.00 & 1.87 & 0.91 & 1.54 & 2.23 & 2.86 & 1.94 \\
    \bottomrule
    \end{tabular}
    \caption{Average of entropy of image-modal probability $P_{img}(y|X)$ and text-modal probability $P_{txt}(y|RC(X))$ according to the number of retrieved texts across 10 datasets.}
    \label{tab:anal1}
    \vspace{-5pt}
\end{table*}

\section{Analysis}
In this section, we present analyses related to confidences from the image-modal and text-modal perspectives.

\subsection{Text Modality Has Lower Entropy than Image Modality}\label{subsec:text_lower}

To examine the characteristics of the image-modal probability $P_{img}$ and text-modal probability $P_{txt}$, we analyze their entropy distributions. Table \ref{tab:anal1} presents the average entropy values of $P_{img}$ and $P_{txt}$ according to the number of retrieved texts. Notably, when only one text is retrieved (i.e., $K$=1), the text modality exhibits significantly lower entropy compared to the image modality. This indicates that the text modality tends to be more overconfident in its predictions compared to the image modality. This phenomenon might be attributed to the shared text encoder between retrieved captions and text labels, where retrieved captions sometimes directly include the text labels.

Therefore, in alignment with a previous study \cite{lakshminarayanan2017simple}, we use multiple captions to prevent overconfident incorrect predictions. It is confirmed that the entropy of the text modality increases as the number of retrieved captions increases. When $K$ is equal to 16, the average entropy becomes most similar between the two modalities, coinciding with the best performance of \texttt{X-MoRe}, shown in Section \ref{subsec:num_cap}.



\subsection{Confidences Should Be Adjusted}
The different entropy distributions between the two modalities, as described in Section \ref{subsec:text_lower}, leads to the different confidence distributions between the two modalities. For instance, while image-modal confidence $\alpha_{img}$ might range from 0.4 to 0.6, the text-modal confidence $\alpha_{txt}$ might range from 0.7 to 0.9. This distortion consistently makes the text modality more trustworthy. Moreover, although the average confidences are similar, a per-sample confidences can differ during test time $t$. To address this issue, we introduced an independent min-max adjustment for each modality. We compare the zero-shot performance between using modal-confidence without adjustment (i.e., $\alpha^t_{img}$ and $\alpha^t_{txt}$) and with adjustment (i.e., $\widehat\alpha^t_{img}$ and $\widehat\alpha^t_{txt}$). Table \ref{tab:anal3} demonstrates the necessity of such independent confidence adjustment.

\begin{table}[h!]
\small
    \centering
    \begin{tabular}{lcccc} \\
    \toprule
    Confidence & Flower102 & Caltech101 & Cars & EuroSAT\\
    \midrule
    ($\alpha^t_{img}$, $\alpha^t_{txt}$) & 69.67 & 90.10 & 60.58 & 32.04  \\
    ($\widehat\alpha^t_{img}$, $\widehat\alpha^t_{txt}$) & \textbf{69.87} & \textbf{91.15} & \textbf{60.66} & \textbf{32.94} \\
    \bottomrule
    \end{tabular}
    \caption{Performance of zero-shot classification according to the confidence adjustment.}
    \label{tab:anal3}
\end{table}

\subsection{Correct Modality Has Higher Confidence}
Finally, to examine whether modal-confidence is effectively adjusted, we divide the results into three cases according to the correctness of the predictions of ($P_{img}$, $P_{txt}$, $P_{ens}$): (1) (correct, incorrect, correct), (2) (incorrect, correct, correct), and (3) (incorrect, incorrect, correct).

Table \ref{tab:anal2} presents the average of $\widehat\alpha_{img}/\widehat\alpha_{txt}$ over all test samples for the three scenarios. A value greater than 1 indicates that the ensemble results prioritize the image modality over the text modality, while a value less than 1 indicates the ensemble results prioritize the text modality over the image modality. As we expected, the image modality is more influential than the text modality when $P_{img}$ is correct and $P_{txt}$ is wrong (i.e., case(1)), and vice versa (i.e., case(2)). Furthermore, the last case (i.e., case(3)) shows that the two modalities have similar power when both are crucial for correct prediction than using single modality.



\begin{table}[h!]
    \centering
    \begin{tabular}{lccccc}
    \toprule
    Case & Flower102  & Caltech101 & Cars & EuroSAT \\
    \midrule
        (1) & 1.32 & 1.36 & 1.31  & 2.12\\
        (2) & 0.81  & 0.79 & 0.82 &  0.70\\
        (3) & 0.93 & 0.84 & 0.92  & 0.90 \\
    \bottomrule
    \end{tabular}
    \caption{Average of $\widehat\alpha_{img}/\widehat\alpha_{txt}$ over all test samples for the three scenarios.}
    \label{tab:anal2}
\end{table}

\section{Conclusion}
In this paper, we proposes \texttt{X-More}, a cross-modal retrieval based inference method that maximizes CLIP's zero-shot capabilities from both image and text modalities. Our method involves two key steps: \emph{cross-modal retrieval} and \emph{modal-confidence-based ensemble}. The former includes a coarse level of image-based retrieval which acquires image-text pairs from an external dataset and a finer level of text-based retrieval which directly considers texts given a query image. The latter step involves a probability ensemble between the two modalities, weighted by independent modal-confidence. While our work is currently constrained to scenarios with accessible external datasets, we believe that our approach contributes to the advancement of large-scale multi-modal models.





\newpage
\bibliography{aaai24}

\newpage

\end{document}